\documentclass[fleqn,10pt]{wlscirep}
\usepackage[utf8]{inputenc}
\usepackage[T1]{fontenc}
\usepackage[table]{xcolor}
\usepackage{booktabs}
\usepackage{tabularx}
\usepackage{graphicx} 
\usepackage{subcaption}
\usepackage{float}
\usepackage[section]{placeins}
\usepackage{listings}

\title{The Open Syndrome Definition}

\author[1,2,*]{Ana Paula Gomes Ferreira}
\author[1]{Aleksandar Anžel}
\author[4]{Izabel Marcilio}
\author[5]{Helen Hughes}
\author[5]{Alex J Elliot}
\author[6]{Jude Dzevela Kong}
\author[3]{Madlen Schranz}
\author[3]{Alexander Ullrich}
\author[1,2]{Georges Hattab}
\affil[1]{Center for Artificial Intelligence in Public Health Research, Robert Koch Institute, Berlin, Germany} %
\affil[2]{Department of Mathematics and Computer Science, Freie Universität Berlin, Berlin, Germany} %
\affil[3]{Infectious Disease Epidemiology Department, Robert Koch Institute, Berlin, Germany} %
\affil[4]{Center for Data and Knowledge Integration for Health, Gonçalo Moniz Institute, Fundação Oswaldo Cruz, Salvador, Brazil} %
\affil[5]{Real-time Syndromic Surveillance Team, Field Services, Chief Medical Advisor Group, UK Health Security Agency, Birmingham, UK} %
\affil[6]{Africa-Canada Artificial Intelligence and Data Innovation Consortium, Department of Mathematics and Statistics, York University, Toronto, Ontario, Canada}

\affil[*]{Gomes-FerreiraA@rki.de}

\begin{abstract}
Case definitions are essential for effectively communicating public health threats. However, the absence of a standardized, machine-readable format poses significant challenges to interoperability, epidemiological research, the exchange of qualitative data, and the effective application of computational analysis methods, including artificial intelligence (AI). This complicates comparisons and collaborations across organizations and regions, limits data integration, and hinders technological innovation in public health. 
To address these issues, we propose the first open, machine-readable format for representing case and syndrome definitions. Additionally, we introduce the first comprehensive dataset of standardized case definitions and tools to convert existing human-readable definitions into machine-readable formats. 
We also provide an accessible online platform for browsing, analyzing, and contributing new definitions, available at \url{https://opensyndrome.org}. 
The Open Syndrome Definition format enables consistent, scalable use of case definitions across systems, unlocking AI's potential to strengthen public health preparedness and response. 
The source code for the format can be found at \url{https://github.com/OpenSyndrome/schema} under the MIT license.
\end{abstract}
\begin{document}

\lstset{
    basicstyle=\ttfamily,
    columns=fullflexible,
    keepspaces=true,
    captionpos=b,
}
\renewcommand{\lstlistingname}{Definition}%

\flushbottom
\maketitle
\thispagestyle{empty}

\section{Introduction}

Case definitions are essential tools for public health practitioners. They are used to identify, monitor, and respond to diseases or groups of diseases \cite{bassil2008case, Devlin2004}.
They inform the public, orient public health policies, and guide surveillance indicators, such as syndrome definitions~\cite{Lazarus2001}. 
Developing case definitions requires expert knowledge of the targeted disease(s), consultation of existing definitions, and analysis of available clinical data.
Case definitions are usually written as free-text descriptions of the key characteristics and criteria of the target disease or public health threat.
The goal of a case definition is to provide a consistent description for public health officials, health workers, policymakers, and the general public so they can understand a given threat.
Figure~\ref{fig:measles_case_definition_comparison} shows two different case definitions for the same disease from different provenance~\cite{EUCaseDefinitions2011, IndiaCaseDefinitions}.
The lack of interoperability and standardization of syndromic indicators and case definitions makes it difficult to compare systems and epidemiological situations across different regions and time periods~\cite{sanchezruizSurveillanceSevereAcute2025}.
Free-text definitions, in particular, introduce a fundamental gap in translation between human and machine interpretation, resulting in inconsistent and error-prone automated processing~\cite{li2021neuralnaturallanguageprocessing}. 
Complex logical relationships appear straightforward to trained readers. 
However, they become deeply problematic for computational systems when given in a free-text format. 
Examples of these relationships are conditions and nested criteria.
During outbreaks, these inconsistencies can delay coordinated responses during critical time periods when containing emerging threats is essential.
For example, inconsistent case definitions had far-reaching consequences for global disease surveillance and control efforts during the Coronavirus Disease 2019 (COVID-19) pandemic.
The consequences of these issues include under-reporting and misclassifications of cases~\cite{Koh2020}, compromised accuracy of surveillance data~\cite{Guerin2021,Tsang2020,Badker2021}, compromised resource allocation~\cite{Raveendran2024}, and problems comparing disease burden and intervention effectiveness across countries~\cite{Suthar2021}.
The variability in case definitions and the absence of structured formats to standardize them represent a surprisingly under-explored gap in public health infrastructure.
Moreover, the dearth of structured formats imposes substantial constraints on embracing artificial intelligence (AI) and eventually implementing AI for surveillance and outbreak detection. 
As health systems increasingly rely on computational approaches, a significant challenge to the use of AI as a technology for protecting public health is posed by the gap between narrative definitions and machine-readable formats. 
When disease indicators are monitored in real-time or near real-time for the purpose of early outbreak detection, it is imperative to use a structured format for case definitions. 
This is especially true in the context of syndromic surveillance, where automated data acquisition is employed~\cite{OverviewSyndromicSurveillance}.

\begin{figure}[h]
\centering
\begin{minipage}[t]{0.45\textwidth}
\textbf{ECDC Case Definition}

Any person with fever
AND
\begin{itemize}
\item Maculo-papular rash
\end{itemize}
AND at least one of the following three:
\begin{itemize}
\item Cough
\item Coryza
\item Conjunctivitis
\end{itemize}
\end{minipage}
\hfill
\begin{minipage}[t]{0.45\textwidth}
\textbf{India Case Definition}
\begin{itemize}
\item A suspected measles case is any person with fever and maculopapular rash (non-vesicular)
\end{itemize}
OR
\begin{itemize}
\item any person in whom a health worker or clinician suspects measles infection
\end{itemize}
\end{minipage}
\caption{\textbf{Measles Case Definition Comparison: The definitions from the European Center for Disease Prevention and Control (ECDC)~\cite{EUCaseDefinitions2011} and India~\cite{IndiaCaseDefinitions} are presented on the left and right, respectively}}
\end{figure}
\label{fig:measles_case_definition_comparison}

Earlier studies ~\cite{Krause2006, Ghodsi2017} verify that utilizing a structured format for clinical standards in case definitions leads to enhanced reporting accuracy compared to narrative descriptions. 
These studies emphasize that even well-defined case definitions can be interpreted differently by various users, which can lead to misclassifications, decreased sensitivity, and lower positive predictive value.
A highlight from the related work is when case definitions were successfully used to automate the generation of a new case definition and translate an existing one from medical conditions~\cite{botsisAutomatingCaseDefinitions2013}. 
Notwithstanding this evidence, at the time of writing, no comprehensive effort has been made to develop a standardized, machine-readable format for case definitions.
Adopting this format would eliminate any inconsistencies in interpretation and lay the groundwork for the infrastructure needed to support next-generation public health surveillance systems.

As a matter of fact, clearly structured definitions can help AI systems identify diseases and track outbreaks more effectively. 
This enables machine learning programs to detect cases more accurately, notice subtle changes in disease patterns, and alert us to emerging health threats, even when traditional symptoms are not yet evident.

This work introduces the Open Syndrome Definition (OSD) format, which is open and interoperable and designed for case definitions.
The OSD format offers flexibility and clarity by providing a machine-readable representation of case definitions.
This enables practitioners to create precise, unambiguous descriptions suitable for AI applications and various software tools.
In addition to defining cases, the OSD format incorporates metadata often missing from traditional narrative formats.
This capability enables the reuse of definitions, comparisons with other jurisdictions, version control, and applications in AI/machine learning projects, benchmarks, and publications.
The OSD format complements, rather than replaces, traditional narrative structures. 
It enables computational systems to precisely interpret case definitions, facilitating automated case classification, cross-jurisdictional comparison, and application in machine learning (ML) pipelines.
The format's name, Open Syndrome Definition, reflects its capacity to encompass both case and syndrome definitions.

In conjunction with the OSD format, this work presents the inaugural dataset of its kind: a collection of machine-readable case definitions for a plethora of diseases from a network of countries spanning five continents~---~the Americas, Europe, Africa, Oceania, and Asia.
True interoperability between jurisdictions, reproducible research, and a foundation for more responsive, data-driven approaches to disease surveillance and public health emergency management are all enabled by this common ``language''.

Case definitions are essential tools in public health and epidemiological research. 
However, they often suffer from ambiguity and a lack of standardization. 
This limits their utility for computational processing and interoperability. 
This study addresses the critical gap in translation between human-readable and machine-processable case definitions. 
To accomplish this, the study has three main goals: 
(1) to create the Open Syndrome Definition, a standardized, machine-readable format that preserves the logical complexity of case definitions while eliminating ambiguity, 
(2) to enable the conversion of the machine-readable format to free text and \textit{vice versa}, which is a necessary operational feature, and 
(3) to compile the first extensive dataset of structured case definitions that covers various diseases and jurisdictions. This study aims to support future advances in public health surveillance, syndromic surveillance, and broader epidemiological research by pursuing these goals.

\section{Results}

This section examines the structure and characteristics of the case definition format. It introduces the first case definition dataset and its maintenance tools. Finally, it presents the Open Syndrome Initiative, a collaborative community built around this ecosystem.

\subsection{The Open Syndrome Definition format}

The OSD format was designed to accurately preserve the clinical meaning of case definitions in a machine-readable representation. Examining various case definitions reveals that, despite differences in language and style, they consistently strive to be clear, simple, and concise~\cite{bassil2008case, ToolkitInvestigationResponse2012}. 
We have also deliberately incorporated these essential qualities into our format.

\begin{figure}[!htb]
  \centering
  \includegraphics[width=0.85\textwidth]{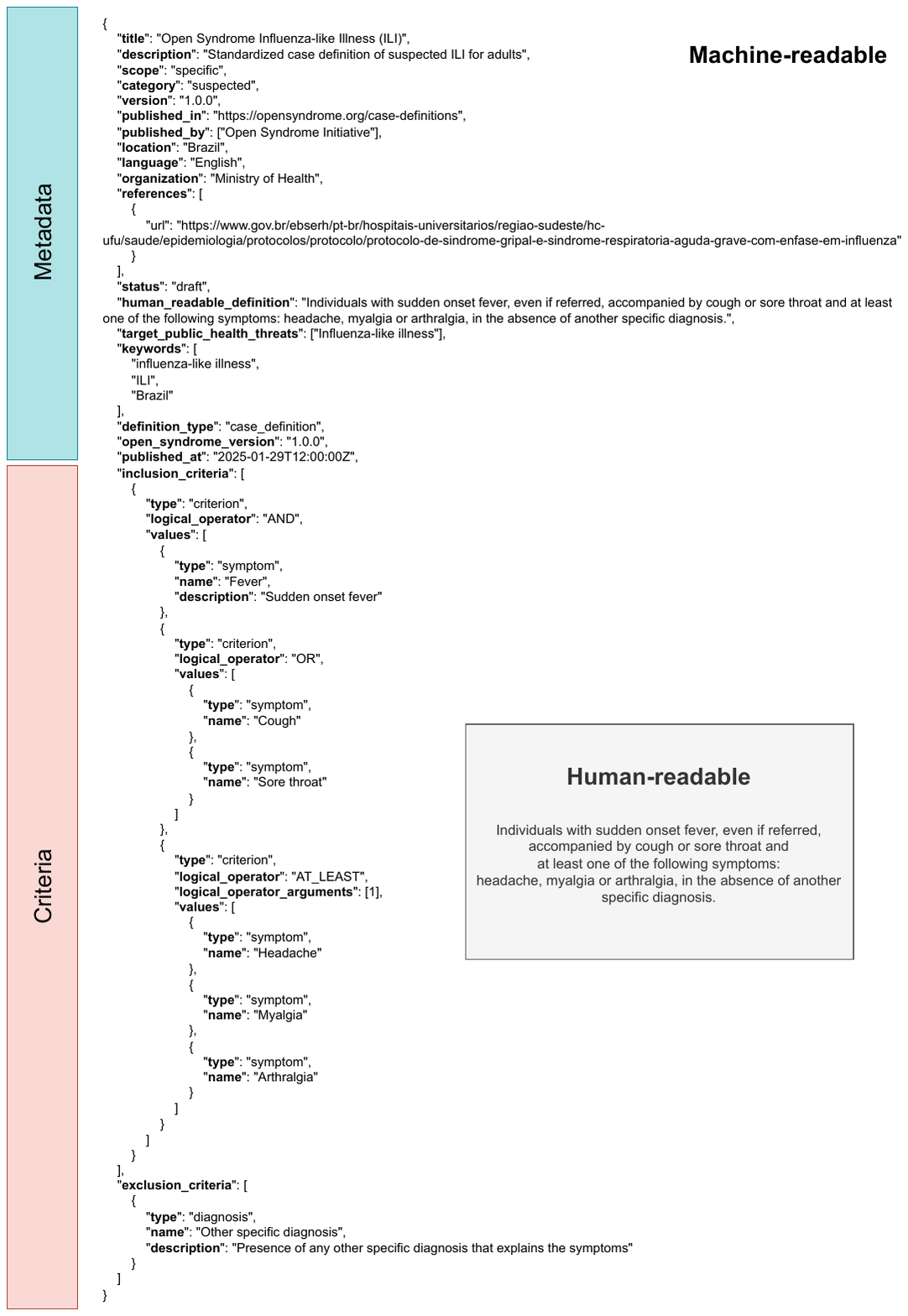}
  \caption{\textbf{
  The human- and machine-readable case definitions of influenza-like illness from the Ministry of Health of Brazil are in the Open Syndrome Definition format.
  The human-readable case definition is shown in the gray box on the right. 
  The machine-readable definition is on the left and uses the JSON format proposed by this work. 
  There are two additional boxes on the sides: the green upper box indicates the metadata fields, and the pink box outlines the inclusion and exclusion criteria.
  }}
  \label{fig:osd_format}
\end{figure}

The OSD format is a JavaScript Object Notation (JSON) schema, a standard for defining the structure and rules of JSON data.
With a JSON schema, one can describe the overall structure of a definition, including its inclusion and exclusion criteria, metadata, properties, and more.
The OSD format transforms traditional narrative case definitions into structured, machine-readable representations.
The version used for the proposed OSD format is draft 2020-12~\cite{JSONSchema202012}.
For example, a free-text definition for Influenza-Like Illness (right side of Figure~\ref{fig:osd_format}) becomes a structured JSON object (left side of Figure~\ref{fig:osd_format}) specifying each criterion with precise logical relationships and the metadata of this definition.

This structured format explicitly defines signs, quantitative attributes, and temporal constraints, which could be ambiguous in a free-text narrative. 
Additionally, it incorporates metadata such as provenance and version history, which is often missing from traditional formats. 
This structured approach enables computational systems to accurately interpret case definitions, thereby facilitating automated case classification, cross-jurisdictional comparison, and application in ML pipelines.
Several core principles guided our design decisions.

\begin{enumerate}[nolistsep]
\item \textbf{Preservation of Clinical Meaning}: Maintaining the integrity of case definitions while keeping a structured format and preserving essential clinical details.
  \item \textbf{AI Readiness}: Designing with computational processing in mind to enable the large-scale analysis and application of artificial intelligence.
  \item \textbf{Interoperability}: Ensuring seamless integration with existing and future systems across different platforms, following the FAIR principles~\cite{Wilkinson2016} and promoting interoperability~\cite{hattab2005theWayForward}.
  \item \textbf{Openness and Accessibility}: Developing a reusable, open format that is free to use and not tied to proprietary platforms.
  \item \textbf{Decentralization}: Allowing for independent implementation across websites, scientific publications, and surveillance systems without central control.
  \item \textbf{Versionability}: Supporting the evolution of definitions over time with clear tracking of changes.
\end{enumerate}

The format addresses several critical challenges in the current landscape. 
These challenges include a lack of standardization in case definitions, potential ambiguity in text-based definitions, fragmented information caused by inconsistent metadata, and barriers to large-scale AI applications resulting from format inconsistencies.
Even though the technical nature of the format might require some knowledge of JSON, we alleviate this by providing user-friendly tools described in Section~\ref{tooling}).

This format has several promising applications, including testing definitions against electronic health record data, creating ML models for automated case detection, comparing disease definitions across countries, and enhancing reproducibility in epidemiological research.

While our collection of data effectively documented a wide array of diseases and methods of definition, we recognize the necessity for subsequent improvements. 
To that end, we included a version field in the format specification. 
This allows us to maintain backward compatibility and accommodate future improvements as the field evolves. 
This versioning approach allows the Open Syndrome Definition format to adapt to emerging needs while preserving access to historical definitions. 
The resulting format balances machine readability with a faithful representation of the original clinical intent. 
This enables automated processing while preserving the essential diagnostic criteria established by public health authorities.

\subsection{Definitions Dataset}
\label{dataset}

\begin{figure}[!ht]
  \centering
  \includegraphics[width=0.75\textwidth]{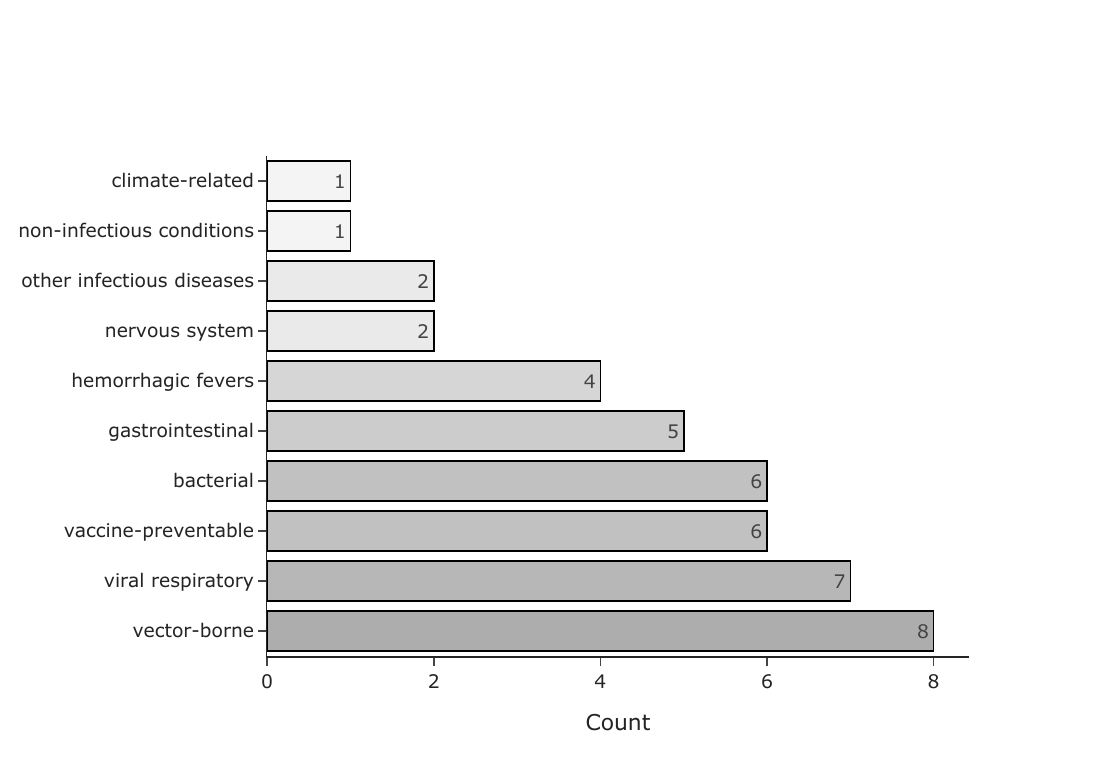}
  \caption{\textbf{Distribution of public health threats categories in the machine-readable dataset definitions} Each condition was classified into a single category based on predominant characteristics, including communicability, pathogen type (viral or bacterial), mode of transmission (\textit{e.g.}, vector-borne), and clinical presentation. Categorization was mutually exclusive, with each case definition assigned to only one group.}
  \label{fig:definition_disease_categories}
\end{figure}

The first comprehensive dataset of machine-readable case definitions was developed, including definitions from 35 countries, three continental organizations, and one global organization.
The dataset contains 40 case definitions for various diseases and is available in three formats:
Open Syndrome Definition JSON, plain text, and the original PDF publications.
The definitions cover infectious diseases, vector-borne illnesses, environmental conditions, and non-communicable diseases. 
Figure~\ref{fig:definition_disease_categories} shows example definitions.

Measles is the most represented disease, with four definitions. Cholera, influenza-like illness (ILI), and COVID-19 each have two definitions. This likely reflects their status as priority conditions in national and international public health surveillance initiatives, in which consistent case definitions are crucial for monitoring morbidity and responding to outbreaks.
The remaining 29 diseases, ranging from common conditions like dengue fever to rare diseases like Lujo hemorrhagic fever, are each represented by one definition.

Additionally, we ensured geographic diversity by selecting at least 10\% of countries from each continent, based on the number of countries represented in the United Nations, to guarantee diversity in geographic representation.
For example, the Pacific Islands are represented collectively through the Pacific Public Health Surveillance Network (PPHSN).
This collaborative effort involves 22 Pacific Island countries and territories that develop case definitions together.
Although using English for our search terms may have introduced some bias, the definitions were available in a range of languages, as depicted in Figure~\ref{fig:continent_language_definitions}.

\begin{figure}
\begin{subfigure}[t]{0.49\textwidth}
\includegraphics[width=\linewidth]{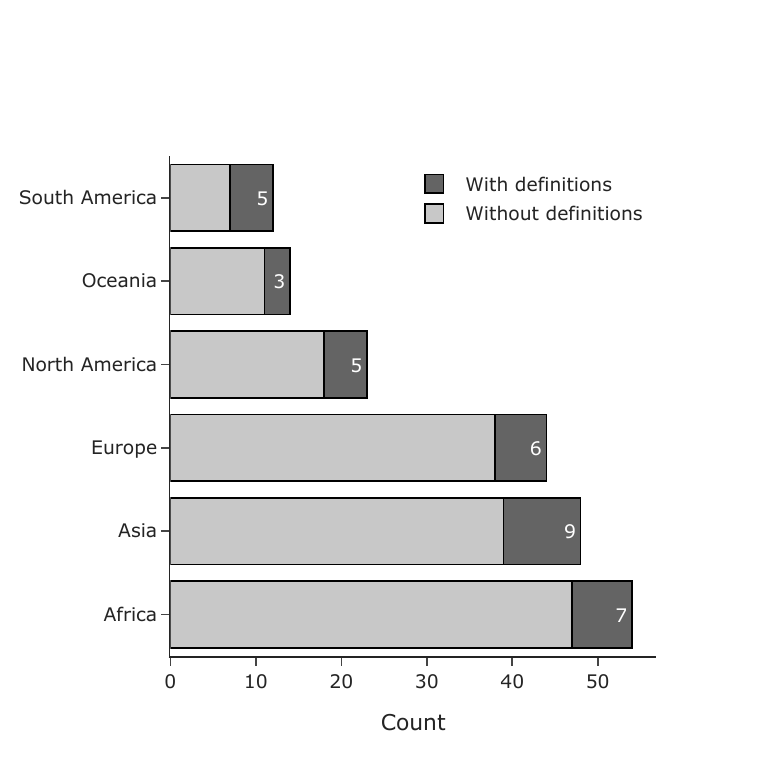}
\caption{Representation of machine-readable definitions by country and continent.}
\end{subfigure}
\hfill
\begin{subfigure}[t]{0.49\textwidth}
\includegraphics[width=\linewidth]{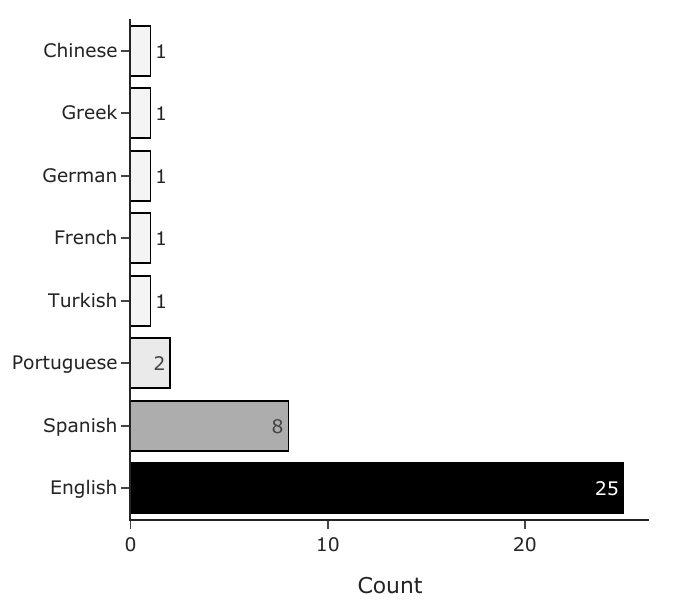}
\caption{Distribution of machine-readable definitions by language.}
\end{subfigure}%
\caption{\textbf{Distribution of definitions by location and language.}}
\end{figure}
\label{fig:continent_language_definitions}

\subsection{Tooling}
\label{tooling}

To overcome barriers to adopting our machine-readable case definition format, we developed a Python command line tool leveraging large language models (LLMs) via Ollama to convert between traditional text definitions and our structured format~\cite{ollama2024}. 
The library offers two main features: it automatically generates structured JSON representations from case definitions and converts JSON definitions back to human-readable text for verification and sharing.

The tool reliably converts between human- and machine-readable formats using locally hosted LLMs. We do not provide a quantitative evaluation of the tool, as this lies beyond the scope of the current work and reflects its auxiliary role in our study. However, our qualitative analysis indicates that Mistral~\cite{jiang2023mistral7b} yields the most reliable results. The tool demonstrates strong fidelity, avoiding hallucinations and accurately translating text into a machine-readable format.
The library was also designed to be flexible, offering two distinct customization options for the conversion process.

First, users can select their preferred LLM model for conversion. Second, users can specify their preferred output language, enabling multilingual text generation, which is essential for the international adoption of public health systems.
By incorporating this flexibility into our library, we ensure local usability and minimize dependencies.

We made the tool available through the Python Package Index (PyPI) at \url{https://pypi.org/project/opensyndrome/} and the project's GitHub repository at \url{https://github.com/OpenSyndrome/open-syndrome-python}, enabling integration into existing workflows. We aimed to bridge the gap between technical and non-technical users by providing extensive documentation on the format, library, and database. 
This makes the Open Syndrome Definition format and its accompanying tools and data accessible to health workers in organizations without dedicated data teams.

\subsection{Open Syndrome Initiative}
\label{initiative}

\begin{figure}[ht]
\centering
\includegraphics[width=0.5\textwidth]{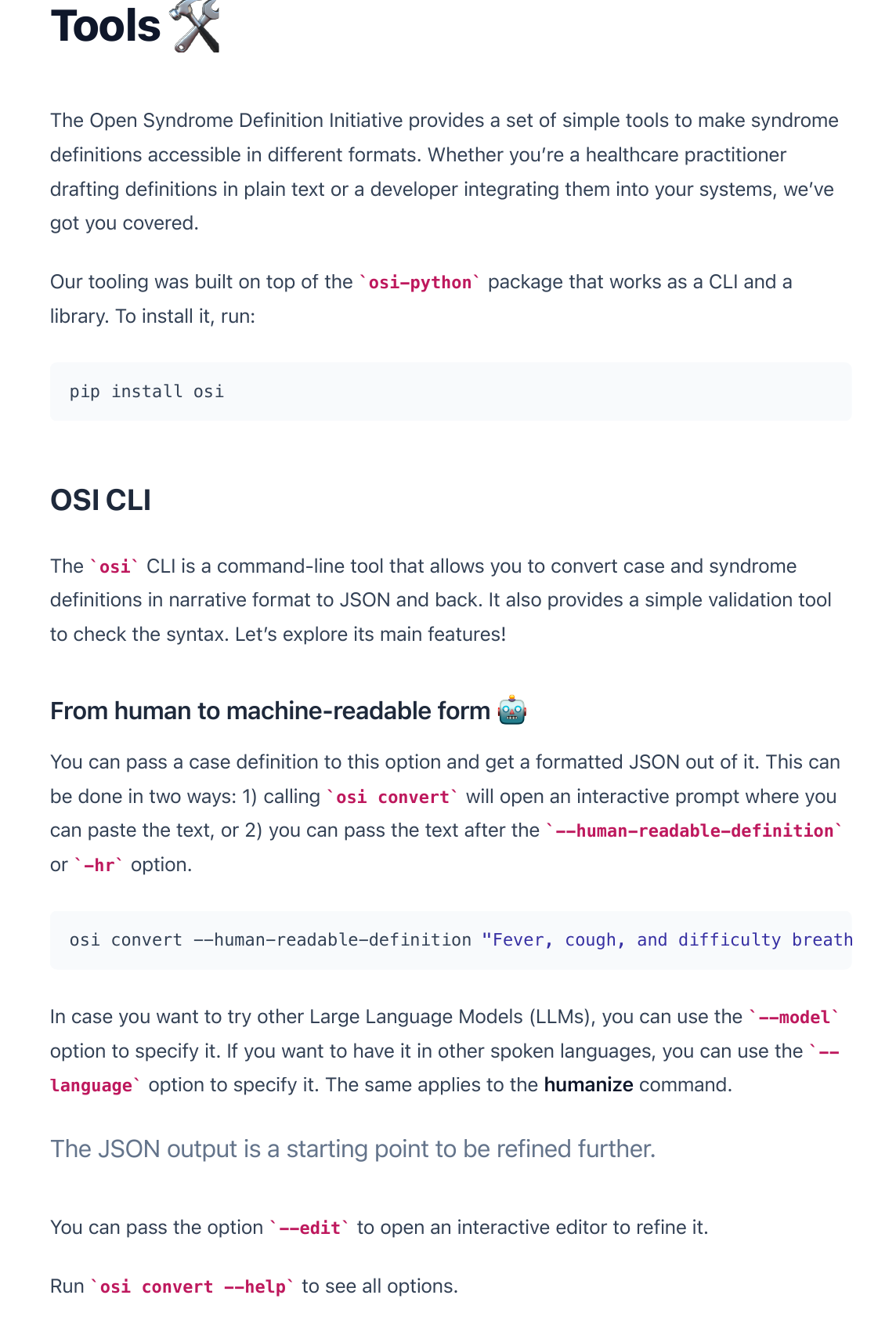}
\caption{Open Syndrome Initiative Tool's page}
\label{fig:tools_documentation}
\end{figure}

The Open Syndrome Initiative (OSI) is a central platform for sharing and accessing case definitions.
The initiative's ecosystem facilitates the practical application of standardized case definitions in public health surveillance. 
It does so by providing a website (\url{https://opensyndrome.org}) where users can search, browse, and analyze definitions. 
The website displays and indexes definitions from the OSI GitHub repository (\url{https://github.com/OpenSyndrome/}), offering the most current versions and statistical information about the definitions. 
Users can contribute definitions or suggest modifications via GitHub pull requests or an online form. This dual approach ensures participation from technical and non-technical users alike.
To ensure high-quality dataset definitions, we implemented a two-stage validation process that combines automated static checks with manual semantic verification upon uploading new definitions. 
The first stage detects missing required fields, formatting inconsistencies, and invalid property types and provides immediate feedback for correction. 
The second stage involves manually reviewing each definition following the automated checks to semantically validate the uploaded data and ensure its accuracy.
The OSI also provides supplementary resources, such as documentation (Figure~\ref{fig:tools_documentation}) and educational blog posts on best practices. 

These resources, alongside the technical infrastructure, support a community of practice around standardized syndrome definitions and enhance global health surveillance capabilities.

\section{Discussion}

This work introduces the Open Syndrome Definition (OSD) format, a novel approach to representing case definitions in machine-readable form. Alongside this format, we presented the first comprehensive dataset of case definitions and developed supporting tools through the Open Syndrome Initiative to foster a community of practitioners. These contributions address a significant gap in public health surveillance infrastructure, as standardized digital representations of case definitions have been notably absent.

Our findings demonstrate that the Open Syndrome Definition format is flexible enough to accommodate case and syndrome definitions in various clinical contexts. The format's nested structure effectively captures the logical relationships present in traditional case definitions. Based on our curated dataset of structured definitions, we found that 53\% of the definitions primarily rely on symptom-based criteria. We assessed the logical complexity of each definition by analyzing the depth of nested Boolean expressions and observed that, although logical groupings were common, nesting rarely exceeded three levels. This suggests that complex clinical conditions can be represented within relatively constrained logical structures.

The development process revealed important insights about case definition structures internationally. The predominance of symptom-based criteria across geographical regions and disease categories indicates a universal approach to case definitions, despite variations in healthcare systems and resources. Additionally, the recurring use of logical operators across definitions from different origins indicates a natural convergence toward structured diagnostic thinking, which our format explicitly codifies.

Despite these promising results, our work has several limitations that warrant consideration. First, while the current dataset is diverse, it is relatively small, containing only 40 case definitions. Although our selective sampling technique covered all continents and major healthcare systems, the limited sample size may not capture the full range of definitions used globally.
We recognize that this initial, one-of-a-kind dataset will serve as a solid foundation for work ultimately developed by and for the community. This is why we developed additional tools to ensure a streamlined, user-friendly expansion process accessible to non-experts.
Additionally, the multilingual nature of our source definitions enriches the dataset's diversity but introduces complexity in accurately translating clinical concepts across language barriers. To address this issue, we have included a versioning field in the format of our open-source repositories. This enables the community to propose modifications or corrections for any errors or omissions.
Furthermore, the lack of established benchmarks or baselines in this field makes comparative evaluation difficult.

Nevertheless, our work establishes the Open Syndrome Definition format at the intersection of significant developments in public health informatics. Its JSON-based structure allows for seamless integration with AI/ML projects, enabling researchers to use these standardized definitions when developing and validating algorithms. As Wang \textit{et al.} noted in their work on automating case identification~\cite{Wang2023}, structured case definitions can significantly improve the efficiency of surveillance systems. Similar applications have demonstrated that standardized case definitions enhance the assessment of information while alleviating the burden on physicians and clinic managers~\cite{Salmaso1997, Radoï2013, Chan2005, WilliamsonE830, Collin2016}.

Our approach's interoperability means that any software capable of parsing JSON can work with these definitions, significantly lowering technical barriers to adoption.
This universal accessibility promotes international collaboration and knowledge sharing, enabling public health officials to quickly adopt and implement definitions from other regions. 
This capability could substantially accelerate the establishment of effective surveillance systems during disease outbreaks, especially in settings with limited resources.

Looking ahead, the Open Syndrome Definition format offers several promising avenues for future development. 
Users will be able to download definitions from various global sources and estimate the number of potential cases they could identify in their local data. 
This feature will speed up the development and adoption of locally optimized definitions and facilitate the implementation of ML models that leverage this standardized format. 
As more institutions contribute to and adopt this approach, we anticipate accelerated collaboration across borders and more responsive surveillance capabilities worldwide.

While our present efforts are concentrated on case definitions, there is also considerable potential to broaden the ready implementation of our methodology to syndromic surveillance. 
The principle of syndromic surveillance is to monitor patients presenting with symptoms, chief complaints, or other non-laboratory/confirmatory diagnoses. 
Public health experts coordinating syndromic surveillance map individual clinical codes to more generic syndromic indicators. While these are not case definitions, as they are purely data- and code-based, similar problems exist with the consistency of syndromic indicators internationally. 
Using OSD to present the code mappings that underpin syndromic indicators and to standardize these indicators internationally would support the global effort to coordinate syndromic surveillance between countries.

The versioning system in our format makes it easy to adapt to changing needs while also ensuring that older versions remain compatible. 
This addresses a crucial requirement for sustainable public health infrastructure. 
As the Open Syndrome Initiative expands its community of practitioners, we anticipate that the collective expertise will further refine the format and available definitions, thereby creating a robust ecosystem of standardized syndromic surveillance resources accessible to all.

A significant future development is improving the AI-driven natural language processing (NLP) techniques that transform narrative definitions into OSD-format representations. The local LLMs of the current system, which are accessed through Ollama, could be supplemented with BioBERT~\cite{lee2020biobert} or ClinicalBERT~\cite{huang2019clinicalbert}. These models are specifically trained on multilingual and multicultural clinical datasets, which could improve parsing precision and flexibility. Furthermore, AI anomaly detection techniques (\textit{e.g.}, isolation forest and transformer-based systems) could be used to automate the validation and correction of definitions, thereby improving the continuous quality of the dataset. AI federated approaches would also enable collaborative algorithm development across jurisdictions while ensuring data confidentiality, which is essential for international disease monitoring and surveillance. Public health federated learning frameworks, such as Flower~\cite{beutel2020flower} and PySyft~\cite{ziller2021pysyft}, could train and evaluate OSD-format-based ML models across regions without sharing sensitive health information. This would improve cross-border analysis, equity, fairness, and public health. OSD format may advance from a defined structure for collaborative model development by progressing in this way.

In conclusion, the Open Syndrome Definition format represents a significant step toward standardizing and digitizing case definitions for public health surveillance. It establishes a foundation for enhanced interoperability, collaboration, and automation in syndromic surveillance. The format's potential applications extend beyond traditional public health contexts into clinical research, healthcare delivery, and emerging disease response.

\section{Methods}

This section describes the methodology used to develop the OSD format, dataset, and supporting tools. We took an iterative approach to the development process. The format evolved through continuous refinement as we collected case definitions, which became our dataset, and created supporting tools. Although this section is organized into distinct subsections for clarity, these components were developed concurrently and informed each other throughout the research process.

\subsection{Definitions Dataset}

Case definitions have specific guidelines regarding writing style. 
These guidelines emphasize simplicity and conciseness and encourage the use of a narrative format~\cite{bassil2008case}. 
Many governmental~\cite{Falldefinitionen, SurveillanceCaseDefinitions2025, governmentofcanadaCaseDefinitionsNationally2000} and public health organizations~\cite{EUCaseDefinitions2011, KRHomepageKnowledge} publicly share their case definitions to ensure that health workers and the public can access information about monitored diseases and their characteristics. 

To identify scientific papers mentioning datasets related to case or syndrome definitions, we searched for the keywords \texttt{case definition + dataset} and \texttt{syndrome definition + dataset} across various scientific sources, as shown in Table~\ref{tab:dataset-search-queries}. 
The choice of terms reflects the core concepts of interest, case definitions, and their association with publicly available or structured datasets. 
Our goal was to cast a broad net without overly constraining the results, so we used simple, inclusive phrases and applied no filters based on time or language. Despite these efforts, however, we were unable to identify any relevant datasets.
\begin{table}[H]
\centering
\begin{tabularx}{\textwidth} { l X X }
\toprule
\textbf{Source} & \textbf{``case definition'' + ``dataset''} & \textbf{``syndrome definition'' + ``dataset''} \\
\midrule
PubMed~\cite{pubmed} & 53 results & 0 results \\
OpenAlex~\cite{openalex} & 213 results & 12 results \\
Kaggle~\cite{kaggle} & 4 results & 0 results \\
HuggingFace~\cite{huggingface} & 2 results & 0 results \\
Harvard Dataverse~\cite{harvard_dataverse} & 14 results & 0 results \\
\textbf{Total} & 286 results & 12 results \\
\hline
Confirmed datasets & 0 & 0 \\
\bottomrule
\end{tabularx}
\caption{\textbf{Search queries for case and syndrome definition datasets accessed on March 24, 2025}. This table summarizes the results of searches conducted in two scientific article databases (PubMed and OpenAlex) and three dataset platforms: Kaggle, Hugging Face, and Harvard Dataverse. The searches yielded 286 results for ``case definition dataset'' and 12 results for ``syndrome definition dataset'', but none of these were actual datasets for case or syndrome definitions.}
\label{tab:dataset-search-queries}
\end{table}

Because our initial search yielded no results, we took a targeted approach to compile case and syndrome definitions for our dataset. Our methodology was a multi-step heuristic process.

First, we focused exclusively on World Health Organization (WHO) member countries.
Second, to ensure balanced representation, we aimed to include countries from each of the following continents: the Americas, Europe, Africa, Oceania, and Asia.
Third, for each country, we began with the search query \texttt{<country name> + case definitions}; if no relevant results were found, we successively tried \texttt{<country name> + syndromic surveillance definitions} and \texttt{<country name> + syndrome definitions}.
Fourth, we aimed to include at least one definition from at least 10\% of the countries within each continent. Once this target was met, we moved on to the next country and then the next continent. When a country provided multiple definitions on a single webpage or PDF, we selected a disease or group of diseases accordingly.
DuckDuckGo was our primary search engine.
For every successful find, we exported the webpage or PDF file featuring case definitions. Since a page or file may contain multiple definitions, we extracted the text to create a machine-readable version.
Among WHO member countries, Japan, Indonesia, Russia, and Cuba had no public definitions.

Ultimately, we collected a total of 40 case definitions. This collection included 36 definitions representing 60 countries, including a group of 22 countries from the Pacific Public Health Surveillance Network (PPHSN); three continental organizations: the Pan American Health Organization (PAHO), the European Center for Disease Prevention and Control (ECDC), and the Africa Centres for Disease Control and Prevention (Africa CDC); and the WHO as a global entity.

As we gathered definitions, common themes surfaced across various diseases and regions, including symptoms, diseases, epidemiological links, laboratory tests, and medical evaluations. 
We also recognized recurring logical patterns in how definitions combined criteria. Operators like \texttt{AND}, \texttt{OR}, and especially \texttt{AT LEAST} were frequently used. The \texttt{AT LEAST} operator called for a specific number of criteria to be met, as illustrated in definitions such as \textit{``fever AND at least two symptoms from: cough, headache, loss of smell, back pain''}.
Through careful analysis of existing definitions and relevant scientific literature~\cite{bassil2008case, Salmaso1997, Chapman2010, Guerin2021}, we pinpointed important metadata elements that, while often found in the broader context of portals and publications, were not consistently included with the definitions themselves.
These insights led to the continuous development of our OSD format, which we cover in detail in Section~\ref{structure}.

The dataset is organized into sections for human and machine readability. 
The human-readable section contains original PDF publications, which may cover multiple diseases, as well as TXT files that contain case definitions converted into our format. 
To ensure consistency, all web-published definitions were exported to PDF. The machine-readable section includes JSON representations of these definitions that have been validated with the Ajv validator~\cite{ajvValidator}. The dataset is available from HuggingFace at \url{https://huggingface.co/datasets/opensyndrome/case-definitions}, and a GitHub repository at \url{https://github.com/OpenSyndrome/definitions}.

\subsection{Schema}
\label{structure}

The OSD format converts traditional narrative case definitions into a structured JSON schema with defined properties and types. These properties were derived through an in-depth review of a wide range of existing case definitions to ensure they reflect common structural and semantic elements found in real-world usage.
When analyzing case definitions of different public health threats from various countries and continents, we identified consistent patterns in information groups. More details are provided in Subsection~\ref{dataset}). We observed that the majority (53\%) relied on symptoms to describe conditions, followed by diagnosis (8\%). Additionally, half of the case definitions included criteria that resembled logical operators (such as \texttt{AND}, \texttt{OR}, and \texttt{AT LEAST}) to group related conditions. 
Based on these observations, we developed a nested structure using a data-driven approach combined with our established principles. 
As illustrated in Figure~\ref{fig:osd_format}, we organized the format into two main groups of information: metadata and criteria.
We describe both groups of information below.

First, the metadata information group provides essential context about the definition itself. This includes version information, scope (broad or sensitive vs. narrow or specific), publication details, the responsible organization, the language used, and other provenance information.
Typically, in narrative formats, this metadata is presented implicitly within published documents or websites (\textit{e.g.,} ECDC case definitions~\cite{EUCaseDefinitions2011}, World Health Organization (WHO) Outbreak Toolkit~\cite{outbreakToolkit}). 
We enabled efficient information retrieval, version tracking, and proper attribution by explicitly structuring this information in a machine-readable format.
Furthermore, the metadata information group consists of metadata properties that capture essential contextual information about the case definition, ranging from basic identification (\texttt{title}, \texttt{description}) to publication details (\texttt{published\_in}, \texttt{published\_at}, \texttt{authors}) and geographical context (\texttt{location}, \texttt{language}). We included properties for tracking the status and version of the case definition itself and the OSD schema. 
Most metadata elements are derived directly from the original case definitions, though two properties~---~\textit{Open Syndrome Version} and \textit{Published by}~---~are specific to the format and initiative, which we discuss further in Section~\ref{initiative}. These and other individual metadata properties and their meaning can be seen in Table~\ref{tab:properties}. 

Second, the inclusion and exclusion criteria specify the conditions that determine whether a case meets the definition or should be excluded.
These conditions, referred to as \textit{criteria}, can be combined using logical operators (\texttt{AND, OR, AT\_LEAST}) to express complex clinical relationships. 
We organized the data into key/value pairs (properties) within the JSON schema, enabling computational systems to interpret simple and compound conditions precisely. This structured approach eliminated ambiguities often present in narrative text~\cite{NewmanGriffis2020} while preserving the clinical intent of the original definitions. 
The criteria properties form the backbone of the format and are structured to capture the logical relationships inherent in case definitions. The format distinguishes between inclusion and exclusion criteria, both built upon our \textit{criterion} meta-type, as described in Table~\ref{tab:criterion}. This structure enables the representation of complex clinical reasoning within a machine-readable framework.

\begin{table}[htb]
\centering
{\rowcolors{1}{}{gray!15}
\begin{tabularx}{\textwidth} { l l l }
\toprule
\textbf{Key component} & \textbf{Property} & \textbf{Description} \\
\midrule
Metadata & Title & Case definition title. \\
Metadata & Description & A detailed description of the case definition. \\
Metadata & Scope & Level of specificity. Options: broad or specific. \\
Metadata & Created at & Date when the definition was created. \\
Metadata & Published in & Source or platform where the definition was published. \\
Metadata & Published at & Date and time in UTC when the definition was published. \\
Metadata & Published by & List of case definition publishers in the Open Syndrome Initiative. \\
Metadata & Authors & List of the case definition authors. \\
Metadata & Location & Geographical location relevant to the schema's application. \\
Metadata & Language & Language in which the definition is written (\textit{e.g.}, English, Spanish). \\
Metadata & Organization & Organization or initiative responsible for the schema maintenance. \\
Metadata & Status & Current status of the definition. Options: draft, published, deprecated. \\
Metadata & Keywords & Keywords related to a definition (\textit{e.g.}, COVID-19, mpox, outbreaks). \\
Metadata & Category & Case definition categories~\cite{bassil2008case}. Options: confirmed, probable, suspected. \\
Metadata & Version & Case definition version to be established by the author.\\
Metadata & Open Syndrome Version & Open Syndrome Definition schema version. Currently: v1.\\
Metadata & Definition Type & Case Definition or Syndromic Indicator.\\
Metadata & Surveillance System Type & Type of surveillance system this definition is.\\
\midrule
Criterion & Criterion (meta-type) &  Set of properties and operators to select a case (details in Table~\ref{tab:criterion}).\\
Criterion & Inclusion criteria (criterion) &  Criterion used to include a case.\\
Criterion & Exclusion criteria (criterion) &  Criterion used to exclude a case.\\
Criterion & References & Scientific references which have supported this definition. \\
\bottomrule
\end{tabularx}}
\caption{\textbf{Breakdown of the format by group of information types.} This table outlines the first-level properties of the Open Syndrome Definition format, dividing them into metadata and criteria. The metadata category contains information about the format and syndrome, including location, organization, and version. The criteria category contains structured information that defines the syndrome itself.}
\label{tab:properties}
\end{table}

To capture the logical nature of the collected definitions, we implemented a flexible attribute-operator-value pattern. 
This pattern can accommodate diverse clinical observations ranging from simple Boolean conditions to complex pattern matching through regular expressions. The format supports numerical comparisons (\texttt{>, >=, <, <=, ==, !=}) and text pattern matching (\textit{e.g.} regular expressions, through \texttt{regex}), enabling precise representation of quantitative thresholds and textual patterns. Please note that logical operators are available at the criteria level (see Section~\ref{structure}). For standardized medical terminology, we included support for coding systems such as the International Statistical Classification of Diseases and Related Health Problems (ICD)~\cite{InternationalClassificationDiseases} and SNOMED Clinical Terms (SNOMED CT)~\cite{OverviewSNOMEDCT} through the code property.

The criterion structure can be defined recursively. The values property enables criteria to be nested within other criteria.
This design allows us to represent complex logical relationships, ranging from simple symptom lists to intricate decision trees with multiple levels of criteria.
With this property structure, we strike a balance between human readability and machine interpretability.
We preserve the clinical logic of narrative case definitions while enabling computational processing and analysis.

\begin{table}[htb]
\centering
{\rowcolors{1}{}{gray!15}
\begin{tabularx}{\textwidth} { p{0.13\textwidth} X X }
\toprule
\textbf{Property} & \textbf{Description} & \textbf{Usage details} \\
\midrule
Type & Type of criterion. Options: \texttt{criteria}, \texttt{syndrome}, \texttt{symptom}, \texttt{diagnosis}, \texttt{diagnostic\_test}, \texttt{professional\_judgment}, \texttt{epidemiological\_history}, \texttt{demographic\_criteria} & The \texttt{type} property is mandatory, followed by \texttt{name} or \texttt{values}. \\
Name & Criterion label. & \\
Description & Detailed description of the criterion. & \\
Logical operator & Keywords that represent a logical operation on criteria. Options: \texttt{AND}, \texttt{OR} and \texttt{AT\_LEAST} & The logical operator \texttt{AT\_LEAST} must be used with the number specified in \texttt{logical\_operator\_arguments}. \\
Logical operator arguments & List of arguments to be passed to the logical operator. & \\
Attribute & The referred attribute \textit{e.g.} body temperature, age, onset. & It is used in composition with \texttt{operator} and \texttt{value}. \\
Value & The reference value for the referred attribute. It could be of any data type because it can represent anything in the real world. & Examples: \texttt{true}, \texttt{37.6}, \textit{abnormal but non-specific bowel gas pattern}. This property is used in composition with \texttt{attribute} and \texttt{value}. \\
Operator & Comparison and matching operators. Options: \texttt{>, >=, <, <=, ==, !=, regex} & It is used in composition with \texttt{operator} and \texttt{attribute}. \\
Regex pattern & Regular expression for evaluation and pattern matching. & It is used in composition with \texttt{operator}. \\
Regex flags & Regular expression flags for extra configuration. & It is used in composition with \texttt{regex\_pattern}.\\
Code & A system-agnostic diagnosis code object that holds \texttt{system}, \texttt{code}, and \texttt{display}. & Useful to represent values from the ICD, SNOMED CT, and others. \\
Values & A list whose types are a \texttt{criterion}. & The criteria items should be unique. \\
\bottomrule
\end{tabularx}}
\caption{\textbf{Criterion Meta-Type Properties}. This table summarizes the properties of a criterion, a fundamental component of a definition.}
\label{tab:criterion}
\end{table}

\subsection{Open Syndrome Definition Format}

Our schema was developed iteratively, resulting in the creation of the Open Syndrome Definition format. We designed the format to include both inclusion and exclusion criteria, reflecting common patterns found in various case definitions. We extracted essential metadata components, such as location, publication date, title, authors, and citation details, from the publishing websites and accompanying papers. This framework for metadata ensures proper attribution and provides contextual information for each definition.

We manually adapted the case definition texts into JSON format and used validation tools to verify their structural integrity and compliance. Transforming the texts from human-readable to machine-readable format revealed additional patterns and edge cases, which informed iterative refinements to the format. Throughout this iterative development process, we prioritized flexibility while maintaining structural consistency. This allowed the format to represent definitions across varied health systems, geographical regions, and clinical contexts.

\subsection{Tooling}

As we developed the Open Syndrome Definition format and dataset, we realized that supporting tools were necessary to facilitate the conversion of human-readable case definitions into machine-readable formats. We developed a Python-based toolkit to streamline this process, enabling researchers and public health professionals to efficiently translate traditional text-based case definitions into structured JSON representations, and \textit{vice versa}. The toolkit used Python version 3.11.

We implemented a conversion utility that leverages large language models (LLMs) through Ollama~\cite{ollama2024} (version 0.9.6) for local deployment. The system requires users to install Ollama locally, providing access to various models according to their computational resources and preferences. We systematically evaluated multiple models including llama-3.2~\cite{touvron2023llamaopenefficientfoundation}, mistral-7b~\cite{jiang2023mistral7b}, and deepseek-r1 (both 7b and 8b variants)~\cite{deepseekai2025deepseekr1incentivizingreasoningcapability}, medllama2~\cite{Medllama2}, and qwen2.5-coder~\cite{hui2024qwen2}. We evaluated the performance of the models based on their ability to comprehend medical terminology accurately and remain faithful to the original clinical meaning. Of the models evaluated, llama-3.2, mistral-7b, and deepseek-r1 demonstrated adequate performance for the conversion task. Mistral-7b was selected as the recommended default due to its optimal balance of accuracy and resource requirements.

Additionally, we developed a reverse conversion function that transformed machine-readable JSON syndrome definitions into human-readable formats that support multiple languages. This bidirectional conversion capability ensures accessibility for diverse user groups and facilitates international collaboration in syndromic surveillance systems.

\subsection{Open Syndrome Initiative}

\begin{figure}[!htb]
\centering
\includegraphics[width=\textwidth]{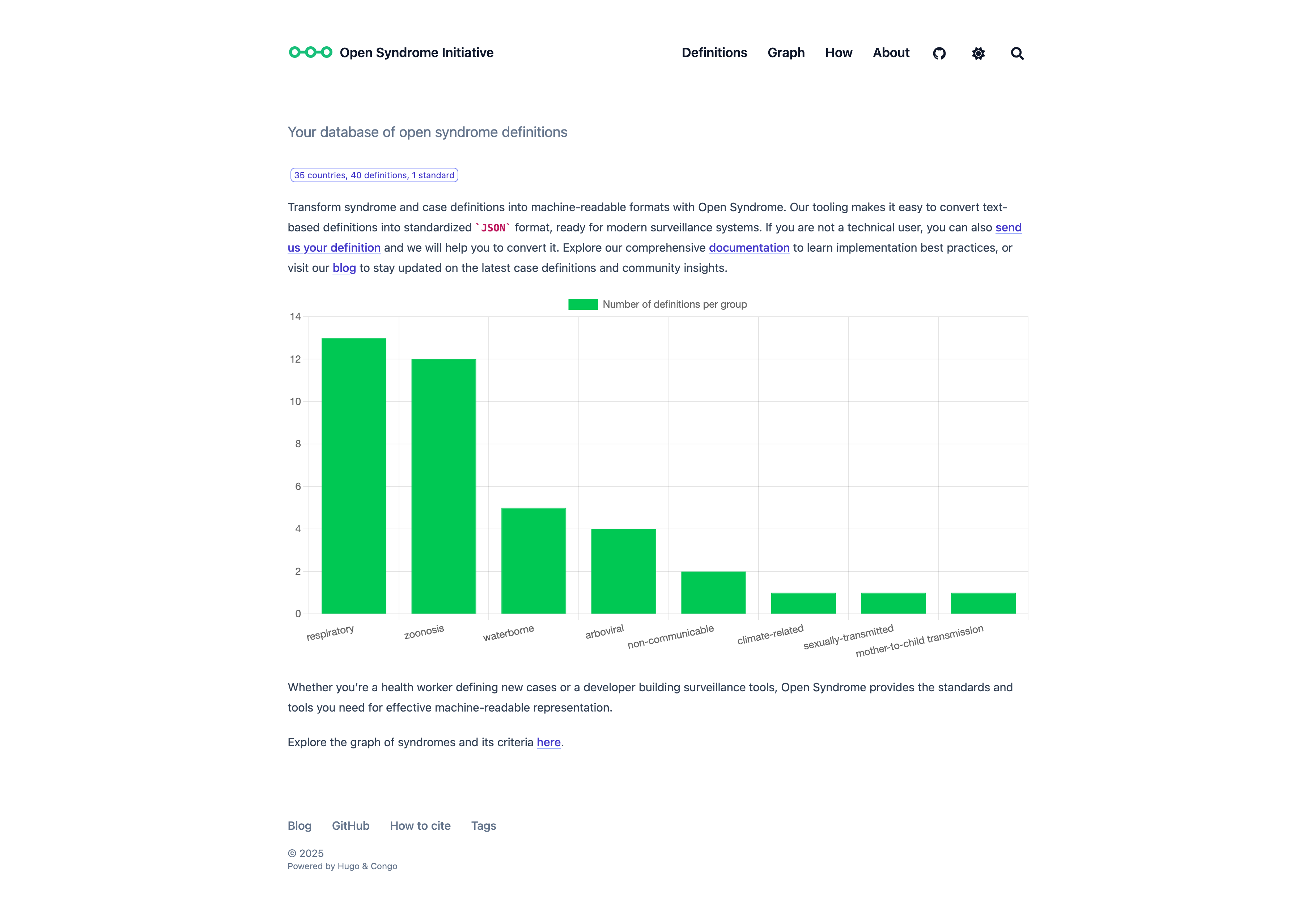}
\caption{\textbf{Open Syndrome Initiative website homepage}}
\label{fig:website}
\end{figure}

To promote the adoption and ongoing development of the OSD format, we created the Open Syndrome Initiative (OSI)~---~a collaborative community platform.
The OSI serves as the central hub for all OSD-related resources and provides an infrastructure for sharing, collaboration, and knowledge exchange. The OSI landing page is presented in Figure~\ref{fig:website}. We cover the methodology behind the website and its functionalities in the following text.

First, to ensure transparency and promote collaboration, we developed the OSI website using open-source tools and libraries. We use Hugo~\cite{hugo} (version v0.145.0) to create the static website which is build using Markdown~\cite{Markdown2025} files. This allows other community members to easily improve or expand the website content with new blog posts or tutorials.

\begin{figure}[!h]
\centering
\includegraphics[width=0.75\textwidth]{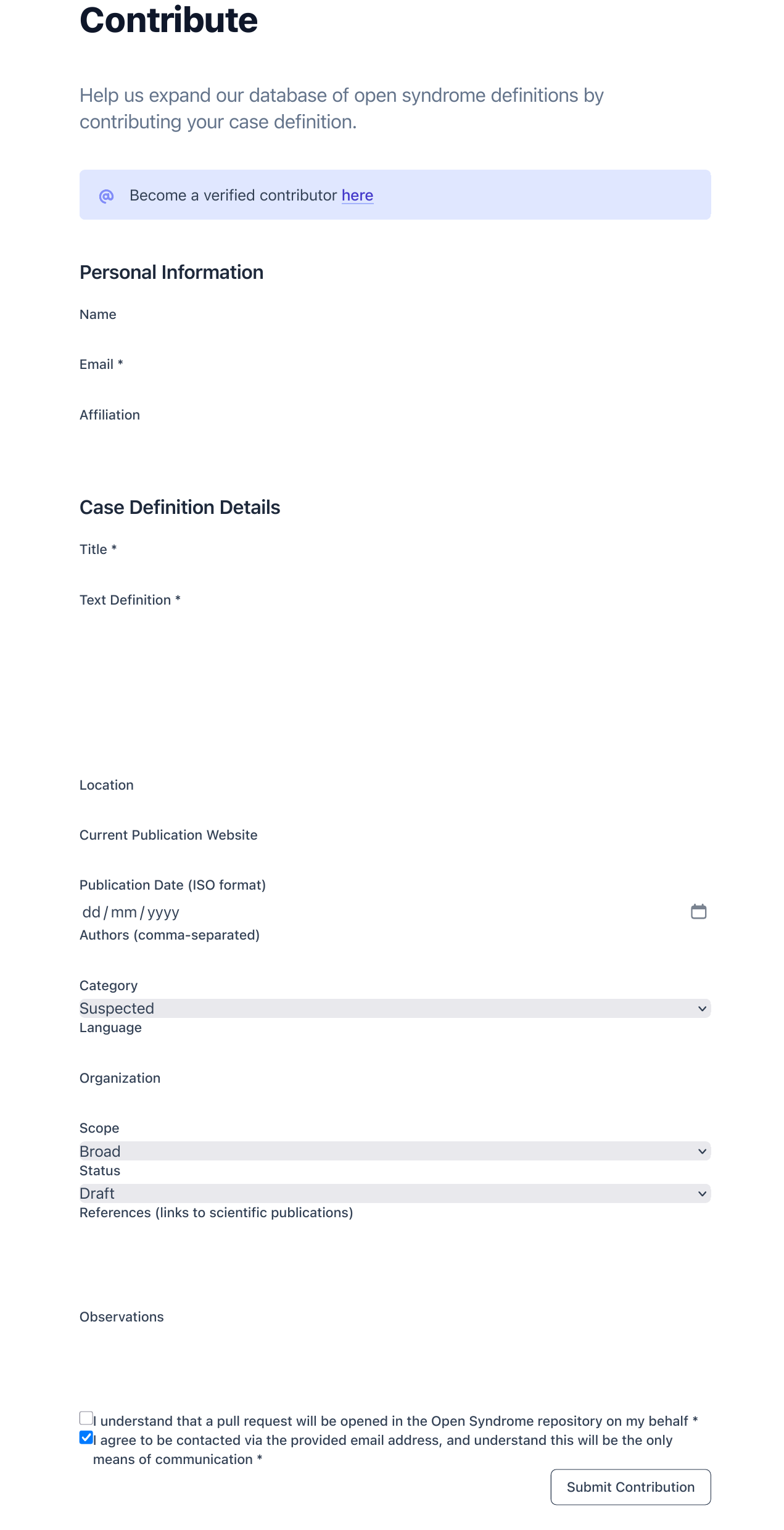}
\caption{\textbf{Open Syndrome Initiative contribution form}}
\label{fig:contribution_form}
\end{figure}

\begin{figure}[!h]
\centering
\includegraphics[width=0.75\textwidth]{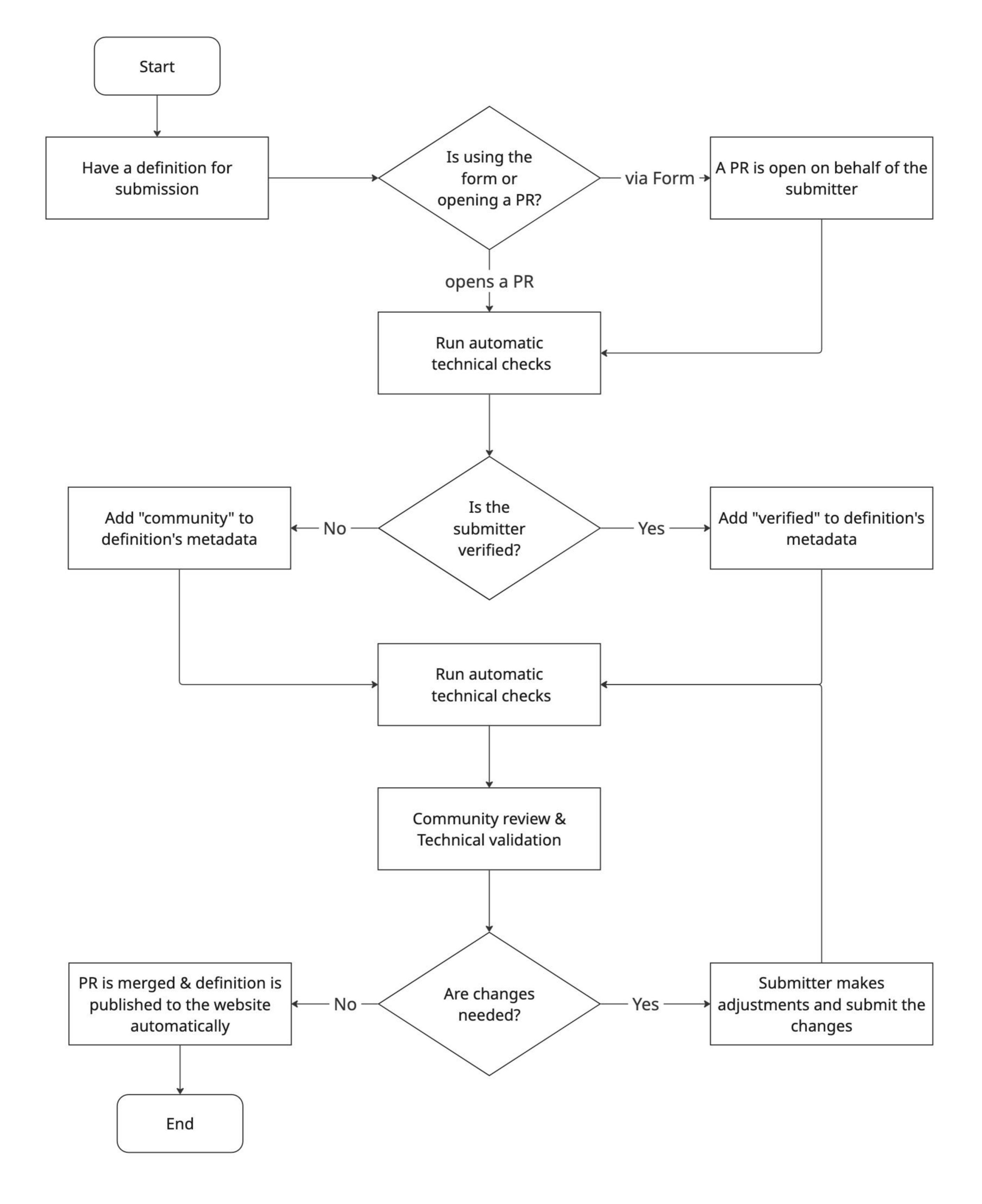}
\caption{\textbf{Open Syndrome Initiative contribution workflow}}
\label{fig:contribution_workflow}
\end{figure}

Second, the website provides a user-friendly contribution workflow for submitting new definitions. This functionality is enabled through pull requests or a simplified web interface\footnotemark (see Figure~\ref{fig:contribution_form}), as depicted in Figure~\ref{fig:contribution_workflow}. All contributions undergo a community review and are published on our website after receiving approval and technical validation. When a definition is submitted via the form, a GitHub Pull Request is automatically created in our repository. This process helps us maintain a clear history of contributions and ensures that all definitions are version-controlled. The submission then follows the same process as any other pull request, including validation and community review. We developed the entire toolkit as open-source software and hosted it on GitHub to enable collaborative development and continuous improvement. To ensure quality and authenticity, we verify the institutional affiliations of first-time contributors.

We implemented the automated validation workflows using GitHub Actions to maintain quality and consistency. These workflows verify JSON schema compliance for each definition and validate new submissions against established formatting requirements. The workflow verify the conformity of submitted JSON files with a predefined schema. This process ensures that the JSON structure is well-formed, that all required fields are present and correctly named according to the schema specifications, and that the values adhere to the expected data formats (\textit{e.g.}, URLs, dates).
We open-sourced the entire toolkit and hosted it on GitHub to enable community contributions and continuous improvement through collaborative development. Our workflow is straightforward: we use a GitHub repository as the central hub where all definitions are stored and managed. Anyone can submit a new definition through the submission contact form or directly via a GitHub pull request. Additionally, individuals can verify their affiliations and become recognized contributors. To do so, they must provide a valid email address and their organization's name through the verification form. Definitions submitted by verified contributors receive a \textit{verified} tag.

Third, the website features an interactive graph visualization of the definitions dataset that we developed using D3~\cite{D3ObservableJavaScript} (version v7.9.0), an open-source JavaScript library for visualizing data, and D3 force-directed graph layout~\cite{D3D3force2025} using velocity Verlet integration.
This feature allows users to explore the dataset intuitively by panning, zooming, and viewing tooltips that display important information about each criterion. The visualization automatically updates to reflect changes in the underlying dataset, so users always see the most current information.

Fourth and last, the web interface includes extra documentation with instructions and suggestions on how to read, work with, and upload case definitions. For those unfamiliar with the format, we provide extensive instructions on how to convert text-based definitions to the OSI format using our tooling.
\footnotetext{\url{https://pypi.org/project/opensyndrome/}}
\footnotetext{\url{https://github.com/OpenSyndrome/}}
\footnotetext{\url{https://opensyndrome.org}}

\section{Conclusion}

The Open Syndrome Definition provides a flexible, interoperable, and machine-readable representation of case definitions. This innovation enables public health professionals, researchers, and technically proficient individuals to exchange epidemiological information more efficiently and consistently. It also enables the broader application of AI and machine learning techniques to public health data.
Along with the format, this work introduces the first dataset of case definitions available in both human- and machine-readable formats. The dataset demonstrates the adaptability of the OSD format across various diseases and geographic settings, paving the way for global comparative analyses of case definition methodologies.
To promote collaboration, we have launched the Open Syndrome Initiative, a platform where users can share definitions and access tools for converting between human- and machine-readable formats.
Public health is inherently collaborative, and the OSD format contributes to this shared effort by promoting better data and more effective tools for preparing for and preventing public health threats.

\section{Author contributions statement}

\textbf{Ana Paula Gomes Ferreira}: Conceptualization, Data Curation, Formal Analysis, Investigation, Methodology, Validation, Visualization, Writing - Original Draft.  
\textbf{Aleksandar An\v{z}el}: Conceptualization, Visualization, Writing - Review \& Editing.
\textbf{Izabel Marcilio}: Discussion, Review \& Editing.
\textbf{Helen Hughes}: Discussion \& Review.
\textbf{Alex J Elliot}: Discussion, Review, Writing \& Editing.
\textbf{Jude Dzevela Kong}: Discussion, Review, Writing \& Editing.
\textbf{Madlen Schranz}: Discussion, Review \& Editing.
\textbf{Alexander Ullrich}: Discussion, Review \& Editing.
\textbf{Georges Hattab}: Conceptualization, Investigation, Supervision, Writing - Review \& Editing.

\section{Additional information}

\begin{description}[align=left]
  \item[\textbf{Acknowledgments}] 
  Alex J. Elliot is affiliated with the National Institute for Health and Care Research (NIHR) Health Protection Research Unit (HPRU) in Emergency Preparedness and Response at the University of Birmingham, as well as the NIHR HPRU in Gastrointestinal Infections at the University of East Anglia. The views expressed in this article are those of the authors and do not necessarily reflect the views of the NIHR, UKHSA, or the Department of Health and Social Care.

  \item[\textbf{Competing interests}] 
  The author declares no conflict of interest.

  \item[\textbf{Data Availability}] 
  All resources associated with this work are publicly available. The schema, case definitions, official website (\url{https://opensyndrome.org}), and the Python package are hosted under the Open Syndrome Initiative GitHub organization (\url{https://github.com/OpenSyndrome}).
 
  The dataset of definitions, including JSON, TXT, and PDF files, is available on Hugging Face (\url{https://huggingface.co/datasets/opensyndrome/case-definitions}). The Python package can also be accessed via PyPI (\url{https://pypi.org/project/opensyndrome/}).

\end{description}

\bibliography{references}

\end{document}